\documentclass[letterpaper, 10 pt, conference]{ieeeconf}  

\IEEEoverridecommandlockouts                             

\usepackage{amssymb}  
\usepackage{amsmath} 
\usepackage{graphicx}
\usepackage{subcaption}
\usepackage{hyperref}

\title{\LARGE \bf
Reinforcement Learning Enabled Adaptive Multi-Task Control for Bipedal Soccer Robots
}

\author{Yulai Zhang$^{1}$, Yinrong Zhang$^{1}$,Ting Wu$^{1}$, and Linqi Ye$^{1}$% <-this % stops a space
\thanks{$^{1}$All authors are with the School of Future Technology, Shanghai University, Shanghai, China.
    {\tt\small yelinqi@shu.edu.cn}
    }
}

\begin{document}

\maketitle
\thispagestyle{empty}
\pagestyle{empty}

%%%%%%%%%%%%%%%%%%%%%%%%%%%%%%%%%%%%%%%%%%%%%%%%%%%%%%%%%%%%%%%%%%%%%%%%%%%%%%%%
\begin{abstract}

Developing bipedal football robots in dynamic combat environments presents challenges related to motion stability and deep coupling of multiple tasks, as well as control switching issues between different states such as upright walking and fall recovery. To address these problems, this paper proposes a modular reinforcement learning (RL) framework for achieving adaptive multi-task control. Firstly, this framework combines an open-loop feedforward oscillator with a reinforcement learning-based feedback residual strategy, effectively separating the generation of basic gaits from complex football actions. Secondly, a posture-driven state machine is introduced, clearly switching between the ball seeking and kicking network (BSKN) and the fall recovery network (FRN), fundamentally preventing state interference. The FRN is efficiently trained through a progressive force attenuation curriculum learning strategy. The architecture was verified in Unity simulations of bipedal robots, demonstrating excellent spatial adaptability-reliably finding and kicking the ball even in restricted corner scenarios-and rapid autonomous fall recovery (with an average recovery time of 0.715 seconds). This ensures seamless and stable operation in complex multi-task environments. The source code and demonstration video are available at
\url{https://openatom.tech/zhang117182/2025competition_gewu_nanoloong}.

\end{abstract}

%%%%%%%%%%%%%%%%%%%%%%%%%%%%%%%%%%%%%%%%%%%%%%%%%%%%%%%%%%%%%%%%%%%%%%%%%%%%%%%%
\section{INTRODUCTION}

As artificial intelligence advances toward general intelligence, embodied intelligence has emerged as a pivotal frontier emphasizing physical embodiment and interaction. Unlike passive perception agents, embodied agents actively interact with physical environments to enable adaptive learning via dynamic sensorimotor coupling, bridging theory with real-world deployment \cite{c1}. Robot soccer—requiring the integration of perception, motion control, and decision-making in dynamic adversarial settings—serves as a critical benchmark for embodied intelligence \cite{c2}. Methods combining learning from demonstration and behavior-based control enhance skill learning efficiency \cite{c3}, while posture stabilization strategies address core disturbance rejection in locomotion \cite{c4}, laying a foundation for reliable small-scale robot operations in competition.

The integration of reinforcement learning (RL) and physical simulation has significantly accelerated the development of robot soccer control. Deep RL successfully trains basic adversarial behaviors \cite{c5}, while skill primitives validate modular collaboration in open-source platforms \cite{c6, c7}. For complex task decomposition \cite{c8}, researchers increasingly adopt embodied methods that couple perception with action \cite{c9, c10}. Additionally, imitation learning \cite{c11} and RL \cite{c12} are crucial for bipedal robot locomotion. System stability and task execution capabilities are further enhanced through multi-posture fall recovery \cite{c13}, multi-objective optimization \cite{c14}, and distributed multi-agent coordination \cite{c15, c16}. Recent studies highlight spatial perception \cite{c17}, optimized neural networks for ball control \cite{c18}, and curriculum learning \cite{c19}, all of which can be efficiently developed within platforms like Unity \cite{c20}.

Despite these advances, developing integrated, multi-skill control policies for bipedal robots remains formidable. First, simultaneously learning foundational locomotion and complex interactions (e.g., ball seeking) intrinsically couples low-level stability with high-level objectives, necessitating intricate reward formulations \cite{c21, c22}. Second, orchestrating seamless transitions between distinct dynamic states (e.g., upright walking and fall recovery) without feature space interference is highly demanding for state modeling \cite{c23, c24}. Thus, harmonizing rhythmic gaits with multi-state task execution remains a key pursuit.

To address these challenges, this paper proposes a modular RL framework for dynamic adversarial tasks in bipedal robots. To resolve task coupling, an open-loop feedforward oscillator maintains stepping rhythms while an RL policy outputs residual actions for high-level maneuvers. Furthermore, a posture-driven state-machine toggles between ball seeking and kicking network and a fall recovery network, ensuring feature decoupling during state transitions. Validated in Unity, the architecture demonstrates robust fall recovery and exceptional spatial adaptability—reliably kicking balls in constrained corners, providing an efficient paradigm for multi-task embodied control.

\section{RELATED WORK}

\subsection{Reinforcement Learning in Robot Soccer}

Robot soccer (e.g., RoboCup) serves as a standard benchmark for evaluating highly dynamic locomotion and object interaction. Traditionally, bipedal robots relied on hand-crafted kinematic models, such as Zero-Moment Point, and scripted controllers for basic walking and kicking \cite{c25}. Recently, end-to-end Reinforcement Learning (RL) has successfully synthesized agile soccer skills directly from environmental interactions \cite{c26}. 

However, learning bipedal locomotion alongside complex interaction tasks typically demands extensive reward formulation. To address this, our study integrates an open-loop oscillator to generate basic stepping rhythms, while an RL policy outputs residual actions for specific soccer maneuvers like ball seeking and kicking.

\subsection{Fall Recovery in Bipedal Robots}

Fall recovery is a critical challenge for bipedal robots, and recent studies have focused on applying RL and multi-task learning to enable robots to recover efficiently from falls. Multi-task learning frameworks aim to optimize the robot’s ability to switch between fall recovery and other tasks such as ball kicking and dribbling. Several approaches, such as \cite{c27}, have investigated fall recovery systems using RL. 

\subsection{Curriculum Learning in Robotic Control}

Curriculum learning is an effective strategy for incrementally training complex tasks by gradually increasing task difficulty. In reinforcement learning, it has been successfully applied to teach robots behaviors, starting with simpler tasks and progressing to more complex ones. In fall recovery, curriculum learning reduces assistance as the robot improves its recovery ability. 

For instance, higher force assistance was initially provided and gradually reduced as the robot became more proficient\cite{c28}. This progressive approach stabilizes learning, reduces failure risk, and improves task performance. In our work, we use curriculum learning to adjust recovery forces based on the robot's progress, ensuring smooth and efficient training.

\section{METHODS}

This section presents a multi-task reinforcement learning control framework for biped robot soccer tasks, aiming to develop an autonomous robot soccer system based on reinforcement learning. 

\subsection{Research Goal}

\begin{figure}[thpb]
    \centering
    \parbox{2.6in}{
        \centering 
        \includegraphics[width=\linewidth]{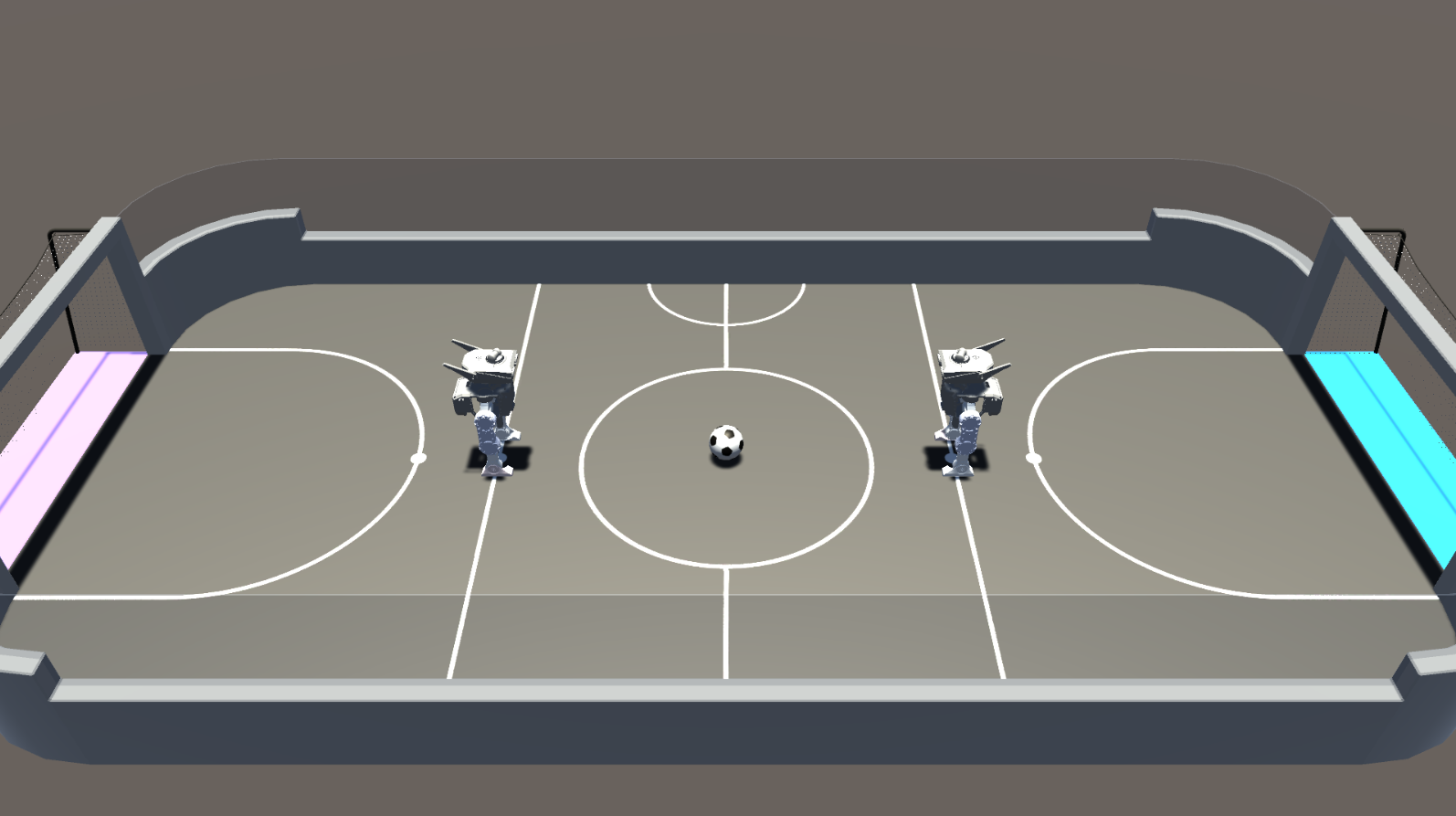} 
    }
    \caption{Soccer Task Scenario.}
    \label{fig:fig4}
\end{figure}

The soccer task scenario is shown in Fig.~\ref{fig:fig4}, which is a sub-module of Gewu Playground built on Unity (https://github.com/loongOpen/Unity-RL-Playground). The scenario includes a soccer field, a soccer ball, and two Tinker robots. The goal is to implement a fully autonomous 1v1 robot soccer match with no human involvement, including the robot’s ability to stand up after falling.

\begin{figure}[thpb]
    \centering
    \parbox{3in}{
        \begin{minipage}{0.21\textwidth}
            \centering
            \includegraphics[width=\linewidth,height=0.18\textheight]{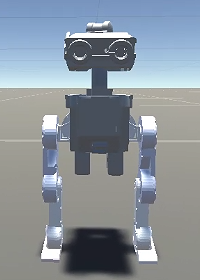}
            \subcaption{Front view} \label{fig:front}
        \end{minipage} \hfill
        \begin{minipage}{0.21\textwidth}
            \centering
            \includegraphics[width=\linewidth,height=0.18\textheight]{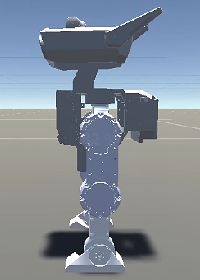}
            \subcaption{Side view} \label{fig:side}
        \end{minipage}
    }
    \caption{Tinker zero position.}
    \label{fig:fig3}
\end{figure}

The robot we used is a small open-source biped robot, NanoLoong-Bipedal (also named Tinker) from OpenLoong (https://github.com/loongOpen/NanoLoong-Bipedal), equipped with 10 rotational joints, with 5 degrees of freedom (DOF) per leg: three at the hip (Y/R/P), one at the knee (P), and one at the ankle (P). Tinker’s zero-position posture is shown in Fig.~\ref{fig:fig3}. This robot is inspired from Disney’s BDX robot and aims to providing an open-source, low-cost research and education platform. 

\subsection{Control Architecture}

\begin{figure}[thpb]
    \centering
    \includegraphics[width=\linewidth,height=0.18\textheight]{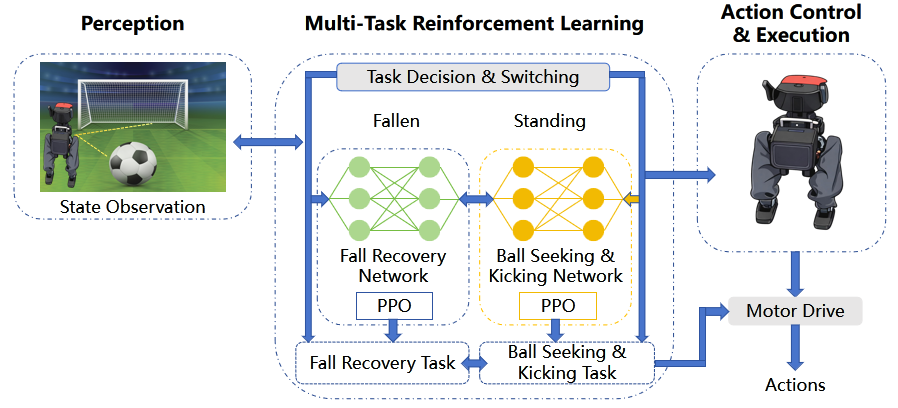} 
    \caption{Multi-Task RL Control Architecture for Tinker.}
    \label{fig:fig1}
\end{figure}

The overall architecture is shown in Fig.~\ref{fig:fig1}.

The framework comprises three core modules: \textit{perception layer}, \textit{multi-task RL layer}, and \textit{action control layer}. The perception layer captures robot posture and soccer environment via state observation for decision-making input. As the core module, the multi-task RL layer dynamically dispatches fall recovery and ball seeking/kicking networks through task decision/switching mechanisms to handle these two core tasks. The action control layer converts decision outputs into motor drive commands for physical action execution.

\subsection{Hybrid Feedforward-Feedback Control}

Learning a stable gait purely from scratch via Reinforcement Learning (RL) often suffers from a high-dimensional exploration space and sub-optimal convergence. To address this, we introduce a time-driven feedforward control module based on a Central Pattern Generator (CPG) mechanism, integrating an open-loop oscillator with an RL-based residual policy.

\subsubsection{Periodic Oscillator}
The feedforward controller provides rhythmic gait priors independently of environmental feedback. A complete gait cycle spans $2T_1$ simulation steps. For a given time step $t_p \in (0, 2T_1]$, the oscillator generates two alternating normalized signals, $u_{f1}$ and $u_{f2}$, to simulate the swing and stance phases. In this paper, $u_{f1}$ is defined as:
\begin{equation}
\label{eq:oscillator}
u_{f1}(t_p) = 
\begin{cases} 
\frac{1 - \cos\left(2\pi \frac{t_p}{T_1}\right)}{2}, & 0 < t_p \leq T_1 \\
0, & T_1 < t_p \leq 2T_1 
\end{cases}
\end{equation}
The signal $u_{f2}(t_p)$ is symmetrically defined with a phase shift of $T_1$.

\subsubsection{Residual Action Superposition}
To combine the nominal gait with RL adaptability, a residual learning approach is employed. The RL network outputs residual actions $a_{rl}^i$. The final target angle $\theta_{target}^i$ sent to the low-level PD controller for active joint $i$ is calculated as:
\begin{equation}
\label{eq:theta_target}
\theta_{target}^i =  d_h^i \cdot u_{f} + d_0^i  + k_i \cdot a_{rl}^i
\end{equation}
where $d_0^i$ is the static base offset, $d_h^i$ is the swing amplitude, $u_{f} \in \{u_{f1}, u_{f2}\}$ is the assigned phase signal, and $k_i$ is the scaling gain. This ensures the oscillator handles the basic rhythmic locomotion while the RL policy focuses on subtle balance and steering compensations.

\subsubsection{Phase Encoding Observation}
To synchronize the RL policy with the feedforward oscillator, the temporal variable $t_p$ is encoded as a continuous harmonic vector and concatenated to the state observation space:
\begin{equation}
\label{eq:phase_encoding}
O_{phase} = \left[ \sin\left(\pi \frac{t_p}{T_1}\right), \cos\left(\pi \frac{t_p}{T_1}\right) \right]
\end{equation}
This prevents temporal discontinuities and provides the neural network with smooth phase information to output synchronized residual actions.

\subsection{Dynamic Switching Mechanism of Dual Networks Based on Posture Perception}

\begin{figure}[thpb]
    \centering
    \includegraphics[width=\linewidth]{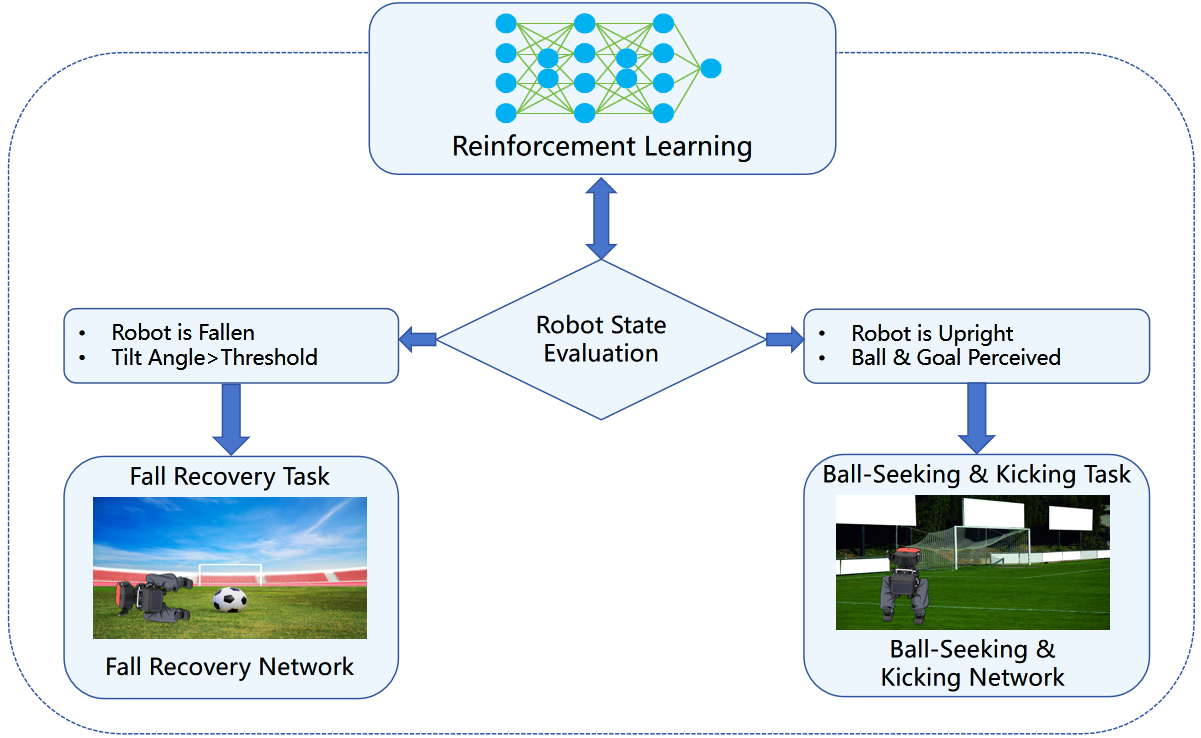} 
    \caption{State Transition Flowchart.}
    \label{fig:fig2}
\end{figure}

To address the inherent conflict between fall recovery and ball seeking and kicking, this framework introduces a posture perception-based dual-network dynamic switching mechanism that allows the Tinker robot to adaptively select the appropriate behavior according to its current posture (Fig.~\ref{fig:fig2}).

\subsubsection{Pose Perception and State Determination} 
Core posture features are collected in real time via Tinker’s IMU and joint encoders, forming a 2D posture feature vector $\mathbf{X}$:
\begin{equation}
\mathbf{X} = [\theta_{tilt} \quad h_{torso}]^T
\end{equation}
where $\theta_{tilt}$ represents the spatial inclination angle between the torso's local upward axis and the global vertical gravity axis; $h_{torso}$ denotes the real-time absolute height of the robot's root link (torso) along the global vertical y-axis. 

A fallen state is robustly identified if the torso inclination angle $\theta_{tilt}$ exceeds $25^{\circ}$ or the torso height $h_{torso}$ drops below a preset physical safety threshold (e.g., $-0.505$\,m in the simulation coordinate frame). If neither condition is met, the robot is considered to be in a stable standing state. This state determination runs at $100$\,Hz (aligned with the physics simulation step of $0.01$\,s) to ensure real-time posture recognition and timely policy switching.

\subsubsection{Dual-Network Architecture and Switching Logic} 
To accommodate the Unity ML-Agents framework dynamically switching models on a single agent, both networks share identical input/output tensor dimensions (a 58-dimensional observation space and a multi-dimensional continuous action space). However, their semantic inputs and reward objectives are strictly decoupled:
\begin{itemize}
    \item Fall Recovery Network (FRN): Focuses exclusively on proprioception. It masks out irrelevant exteroceptive environmental features (e.g., ball and goal positions) with zero-padding. By processing only core posture features and joint states, it outputs residual joint actions specifically optimized for fall recovery and posture stabilization.
    \item Ball Seeking and Kicking Network (BSKN): Utilizes the complete observation vector. It fuses the robot's proprioceptive states with real-time exteroceptive features (relative coordinates of the soccer ball and target goal). It outputs target-driven residual commands to modulate the feedforward gait for ball-seeking, body turning, and kicking.
    \item Switching Logic: Posture perception feeds state data to the task decision module, triggering network switching—FRN activates (BSKN suspends) in fallen states, BSKN reactivates in standing states. A 3-frame continuous judgment buffer mitigates switching jitter: switching occurs only if conditions are met for three consecutive frames, with smooth interpolation on joint action commands to prevent abrupt balance loss.
\end{itemize}

\subsection{Multi-Dimensional Reward Function Design}
\subsubsection{FRN Reward Function Design}

The FRN enables rapid, stable upright recovery post-fall through a two-component reward formulation. The per-step reward is $R = R_{\text{up}} + R_{\text{act}}$, with the components defined as follows:

\begin{itemize}
    \item Upright Reward: \( R_{\text{up}} = \mathbf{u} \cdot \mathbf{v} \): Maximizes the dot product between the torso up-direction unit vector $\mathbf{u}$ and the world vertical unit vector $\mathbf{v}$, approaching $+1$ when upright and negative when tilted or inverted.
    
    \item Actuation Penalty: \( R_{\text{act}} = -2 \) if \( \mathbf{u} \cdot \mathbf{v} < 0.7 \), otherwise \( -0.3 \sum_{i=1}^{N} |u_i| \): Applies a strong constant penalty when fallen, prioritizing recovery. Once upright, the penalty becomes linear in total action magnitude $\sum_{i=1}^{N} |u_i|$, where $u_i$ is the filtered action for joint $i$ and $N$ is the number of controllable joints, encouraging smooth, energy-efficient motions.
\end{itemize}

\subsubsection{BSKN Reward Function Design}

The BSKN targets efficient ball localization, control, and kicking in robot soccer. The total reward is defined as $R_{\text{total}} = R_{\text{progress}} + R_{\text{approach}} + R_{\text{stability}} + R_{\text{efficiency}}$, which incentivizes rapid ball approach, optimal shooting angles, and effective kicking while penalizing energy waste and postural instability. The individual terms are defined as follows:

\begin{itemize}
    \item Forward Reward: $R_{\text{progress}} = \text{forward speed} \times \alpha$: Rewards higher forward linear velocity ($\alpha$ = forward reward coefficient).
    \item Approach Reward: $R_{\text{approach}} = \alpha_{\text{goal}} \times \left(1 - \frac{d_{\text{goal}}}{d_{\text{max}}}\right)$: Rewards goal proximity with sharp growth ($d_{\text{goal}}$ = robot-goal distance, $\alpha_{\text{goal}}$ = reward coefficient).
    \item Postural Stability Penalty: $R_{\text{stability}} = -\beta \times (|\theta_{\text{pitch}}| + |\theta_{\text{roll}}|)$: Penalizes excessive pitch/roll angles to enhance balance ($\beta$ = stability penalty coefficient; $\theta_{\text{pitch}}$, $\theta_{\text{roll}}$ = current pitch/roll angles).
    \item Energy Efficiency Penalty: $R_{\text{efficiency}} = -\gamma \sum_{i=1}^{n} |u_i|$: Penalizes excessive joint energy consumption ($u_i$ = $i$-th joint motion input, $\gamma$ = energy penalty coefficient).
\end{itemize}

Combined, these components guide efficient ball seeking and kicking while minimizing energy use and maintaining postural stability.

\subsection{Parallel Training}

This study deploys 24 parallel instances for RL training on the Gewu Playground platform. This architecture enables simultaneous multi-instance sampling and policy updates, boosting training efficiency/data diversity and reducing overall training time. Each instance independently explores the state space to enrich experience data diversity; cross-instance experience sharing enhances policy stability/convergence, minimizes update variance, and ensures comprehensive coverage of training state distribution.

\section{SIMULATION AND RESULTS}

\subsection{Fall Recovery Network Training Result}

\begin{figure}[thpb]
    \centering 
    \parbox{3in}{
        \centering 
        \includegraphics[width=\linewidth]{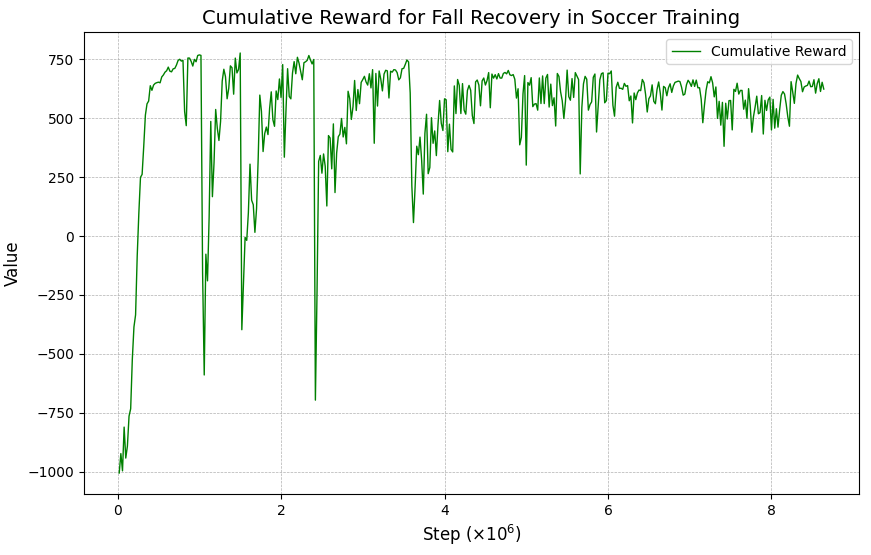} 
    }
    \caption{Cumulative Reward for Fall Recovery Network.}
    \label{fig:fig5}
\end{figure}

As shown in Fig.~\ref{fig:fig5}, the cumulative reward curve of the Fall Recovery Network (FRN) clearly illustrates the effectiveness of the proposed curriculum learning paradigm. Initially, the reward starts at approximately -1000 but ascends sharply to peak near 750 within the first $1 \times 10^6$ steps, facilitated by substantial external assistive forces. The noticeable sharp dips in the learning curve (e.g., around $1 \times 10^6$ , $1.5 \times 10^6$ , and $2.5 \times 10^6$ steps) correspond to the scheduled progressive reduction of these assistive forces. Although increasing the task difficulty causes transient performance drops, the agent rapidly adapts and recovers to high reward levels. After $4 \times 10^6$ steps, as the external assistance diminishes to zero, the curve stabilizes consistently between 500 and 700. 

This process demonstrates that the robot successfully transitions from assisted standing to autonomous fall recovery, thereby validating the proposed reward function and laying a robust foundation for the subsequent dual-network dynamic switching.

\subsection{Ball Seeking and Kicking Network Training Result}

\begin{figure}[thpb]
    \centering
    \includegraphics[width=\linewidth]{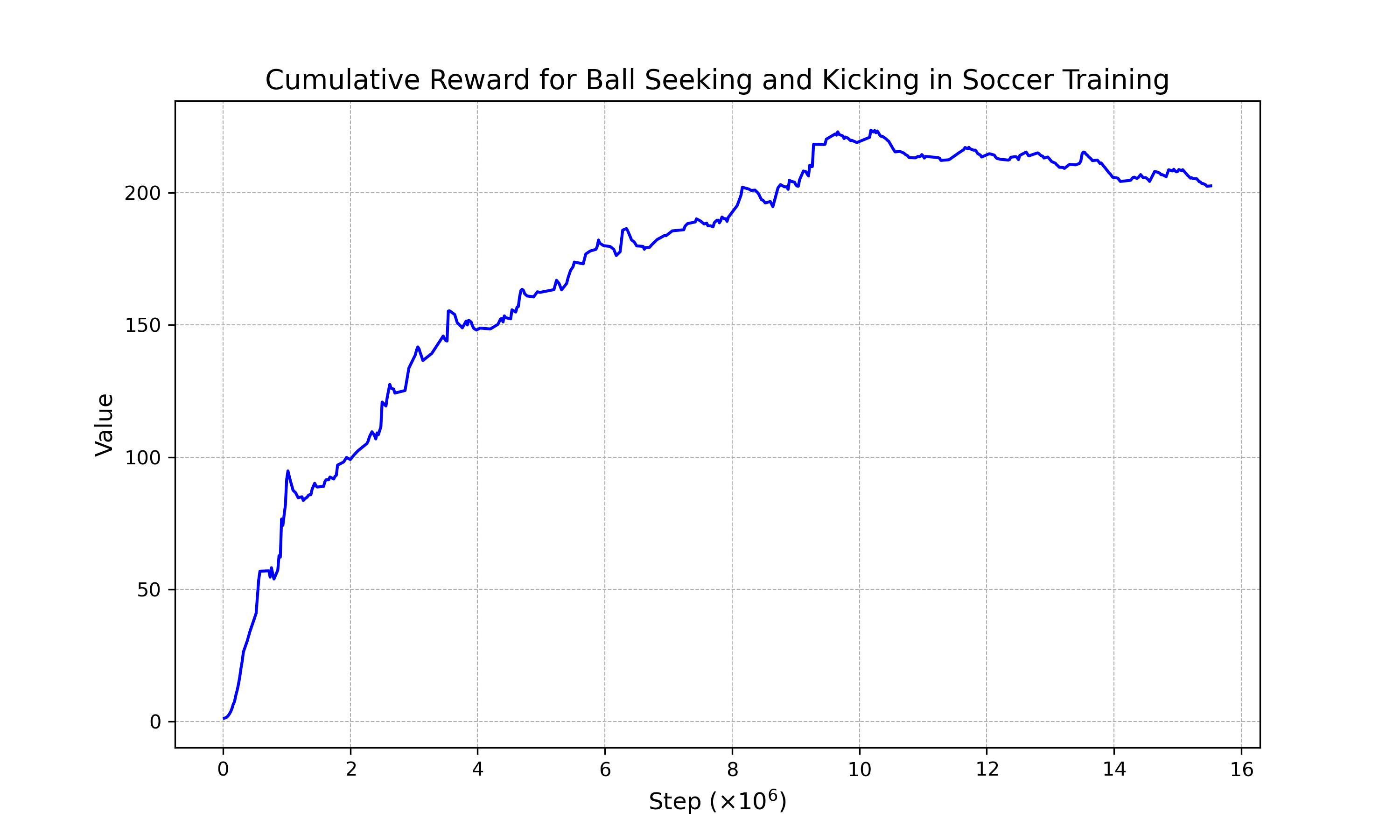} 
    \caption{Cumulative Reward for Ball Seeking and Kicking Network.}
    \label{fig:fig6}
\end{figure}

As shown in Fig.~\ref{fig:fig6}, the BSKN cumulative reward starts at 0, rises sharply to 100 within $3 \times 10^6$ steps (basic ball-seeking/stepping mastery), grows gradually with minor fluctuations (optimal kicking angle exploration), and stabilizes at ~200 by $16 \times 10^6$ steps. This confirms the reward function guides effective learning of the composite ball seeking and kicking task, balancing environmental perception and dynamic motion control.

\subsection{Robot Fall Recovery and Time Distribution Analysis}

To evaluate the fall recovery performance of the robots, a total of 24 instances were tested in a simulated soccer environment. Each instance involved a robot that was intentionally knocked down, and the time taken to return to a stable standing position was recorded. The recovery process was analyzed to assess the effectiveness of the proposed reinforcement learning-based fall recovery system.

\begin{figure}[thpb]
    \centering
    \parbox{3.2in}{
        \centering 
        \includegraphics[width=\linewidth,height=0.10\textheight]{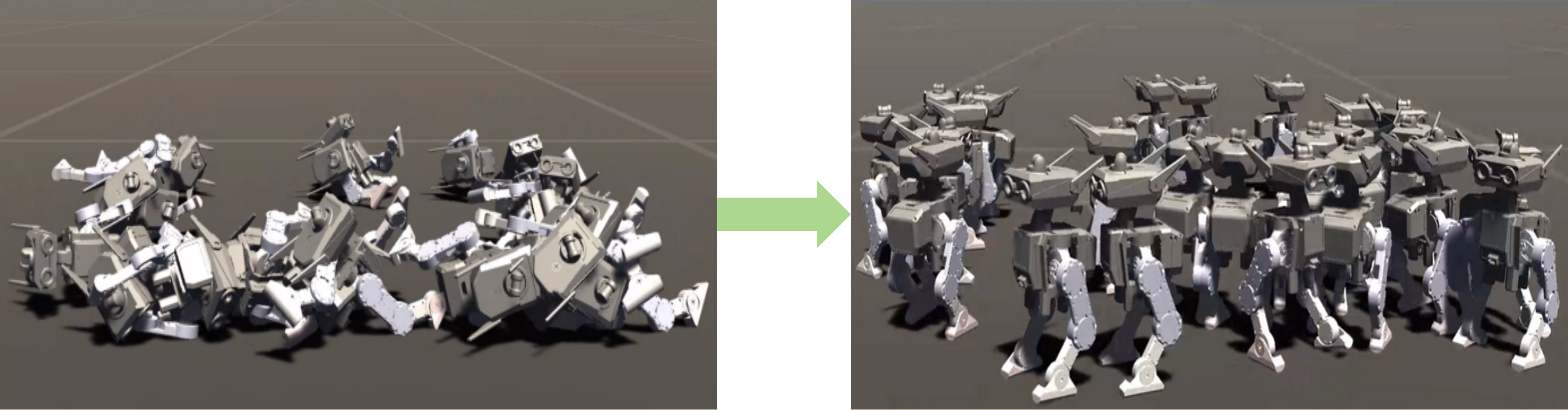} 
    }
    \caption{Robot Fall Recovery Process.}
    \label{fig:Robot Fall Recovery Process}
\end{figure}

Fig.~\ref{fig:Robot Fall Recovery Process} illustrates the dynamic recovery process, showing the transition of the robots from the fallen state (left) to the standing position (right). This transition demonstrates the robot's ability to efficiently recover from a fall and resume its soccer tasks, highlighting the robustness of the proposed system in dynamic environments.

\begin{figure}[thpb]
    \centering
    \parbox{3.2in}{
        \centering 
        \includegraphics[width=\linewidth,height=0.24\textheight]{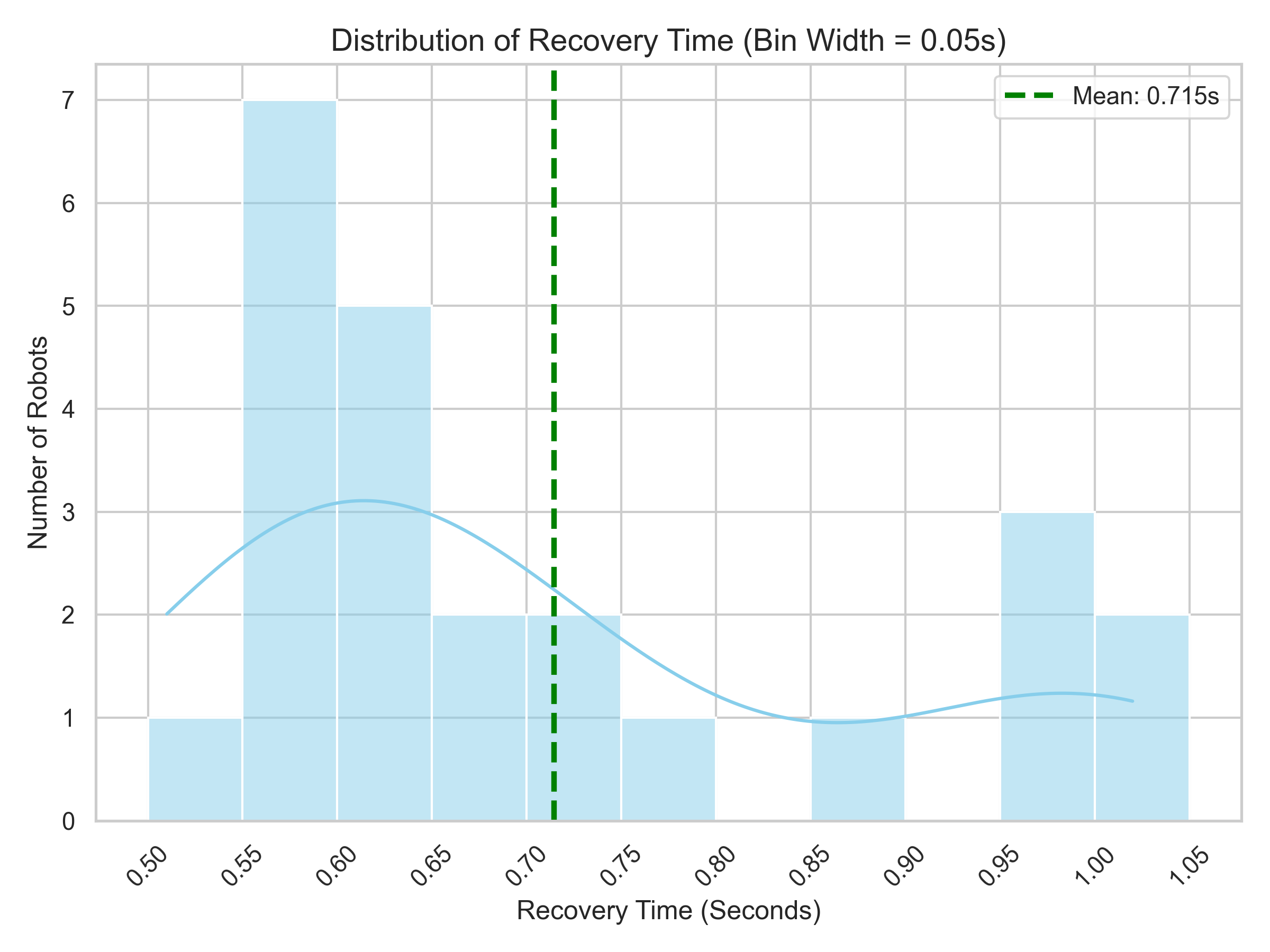} 
    }
    \caption{Distribution of Robot Recovery Time.}
    \label{fig:Distribution of Robot Recovery Time}
\end{figure}

Fig.~\ref{fig:Distribution of Robot Recovery Time} presents the distribution of recovery times across all 24 robot instances. The histogram and the smoothed density curve reveal that recovery times are generally distributed between 0.5 and 1.0 seconds. The majority of robots returned to a standing position within 0.6 to 0.7 seconds, with the mean recovery time marked at 0.715 seconds. The data indicates a stable recovery process, with a few instances requiring slightly longer recovery times. This further supports the robustness and effectiveness of the fall recovery network used in our approach. 

\subsection{Spatial Adaptability in Corner Scenarios}

\begin{figure}[thpb]
    \centering
    \parbox{3.2in}{
        \centering 
        \includegraphics[width=\linewidth,height=0.24\textheight]{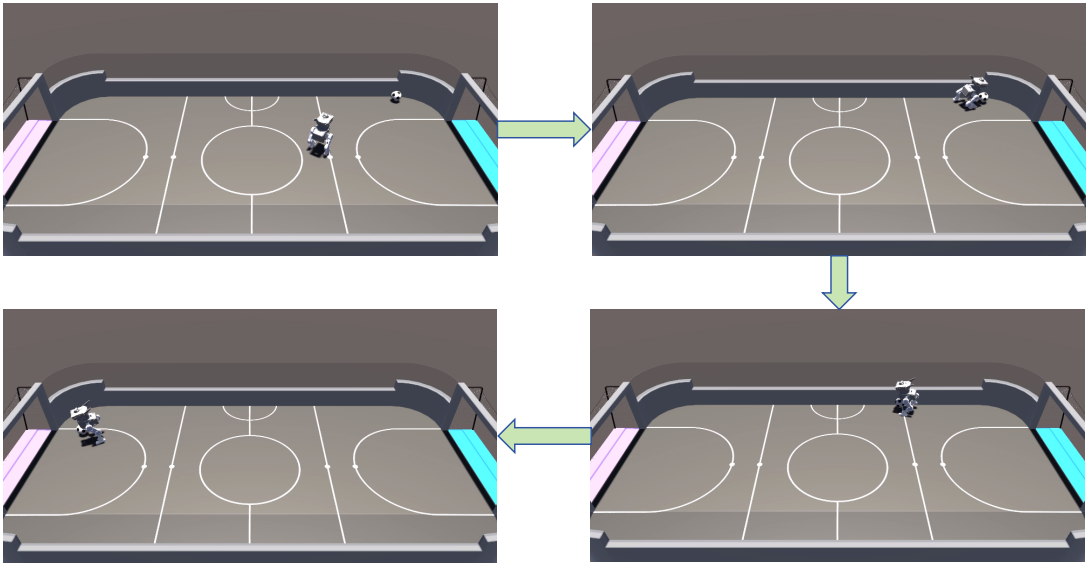} 
    }
    \caption{Spatial Adaptability  in Corner Scenarios.}
    \label{fig:Spatial Adaptability  in Corner Scenarios}
\end{figure}

As shown in Fig.~\ref{fig:Spatial Adaptability  in Corner Scenarios}, the robot demonstrates robust spatial adaptability in a corner scenario. The process consists of four key stages: (a) navigates to the ball in the corner, (b) approaches and aligns its body toward the goal, (c) stabilizes posture for an accurate kick, and (d) executes a successful shot toward the goal. This capability validates that the reward structure, particularly the ball-seeking and target-area components, effectively incentivizes the agent to maintain goal orientation even under challenging spatial constraints.

\subsection{1v1 Soccer Match Results}

To validate the practical applicability of the trained dual-network framework, a functional test was conducted in a simulated soccer scenario. Tinker exhibited smooth, coordinated motion with balanced posture, efficient leg movements for ball tasks, prompt dual-network switching for fall recovery, and seamless resumption of ball tasks post-recovery—confirming stable continuous task execution in complex scenarios and validating training reward curve convergence.

\begin{figure}[thpb]
    \centering
    \parbox{3in}{
        \begin{minipage}{0.21\textwidth}
            \centering
            \includegraphics[width=\linewidth]{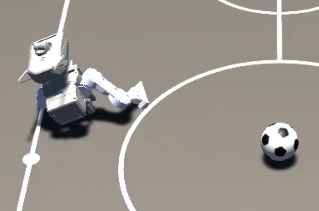}
            \subcaption{Fall during movement} \label{fig:fig7-1}
        \end{minipage} \hfill
        \begin{minipage}{0.21\textwidth}
            \centering
            \includegraphics[width=\linewidth]{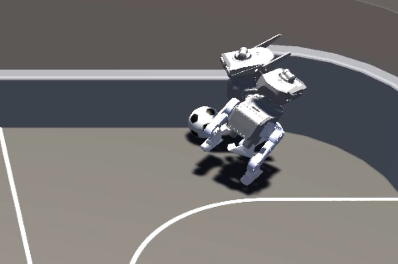}
            \subcaption{Dynamic ball dribbling} \label{fig:fig7-2}
        \end{minipage}
        \begin{minipage}{0.42\textwidth}
            \centering
            \includegraphics[width=\linewidth]{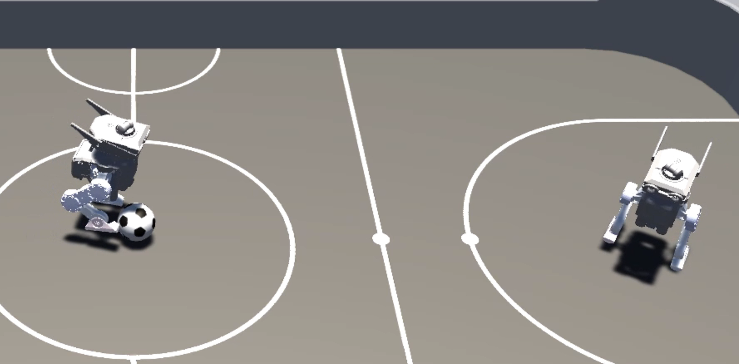}
            \subcaption{Ball seeking/approaching} \label{fig:fig7-3}
        \end{minipage}
        \begin{minipage}{0.21\textwidth}
            \centering
            \includegraphics[width=\linewidth]{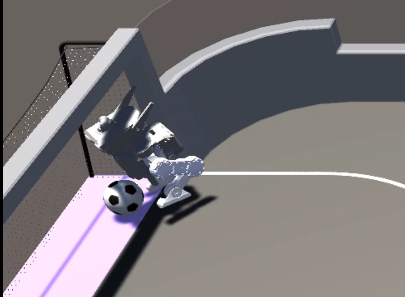}
            \subcaption{Successful goal scoring} \label{fig:fig7-4}
        \end{minipage}
        \begin{minipage}{0.21\textwidth}
            \centering
            \includegraphics[width=\linewidth]{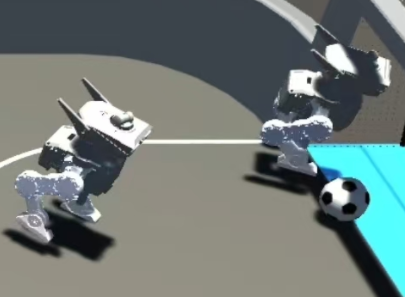}
            \subcaption{Confrontation near goal} \label{fig:fig7-5}
        \end{minipage}
    }
    \caption{Key test scenario snapshots.}
    \label{fig:fig7}
\end{figure}

Key test scenario snapshots (Fig.~\ref{fig:fig7}) visualize core task execution stages: fall-induced network switching (Fig.~\ref{fig:fig7-1}), dynamic dribbling/ball approaching (Fig.~\ref{fig:fig7-2}–\ref{fig:fig7-3}), and full-task-chain execution via goal scoring/near-goal confrontation (Fig.~\ref{fig:fig7-4}–\ref{fig:fig7-5}). These results verify robust posture perception, precise BSKN motion coordination, and effective complex task completion under dynamic adversarial conditions.

\section{CONCLUSIONS}

This paper presented a modular reinforcement learning framework to address the critical challenges of task coupling and state transition in bipedal robot soccer. By employing a hybrid feedforward-feedback control architecture, we explicitly decoupled foundational rhythmic gaits (generated by an open-loop oscillator) from target-driven maneuvers (handled by RL residual policies). Furthermore, a posture-driven switching mechanism dynamically toggles between a primary Ball Seeking and Kicking Network (BSKN) and a dedicated Fall Recovery Network (FRN), with the latter efficiently trained via a progressive force-decay curriculum. 

Extensive Unity simulations validated the robustness of the proposed system. The bipedal robot demonstrated highly agile spatial adaptability in ball manipulation and achieved rapid, autonomous fall recovery with a mean time of 0.715 seconds. Notably, the behavioral transitions were smooth and seamless, effectively eliminating joint jitter. This decoupled control paradigm provides a highly efficient and reliable solution for embodied multi-task agents. Future work will focus on deploying the trained policies to physical hardware via sim-to-real transfer and integrating vision-based perception for multi-agent cooperative strategies.

% \addtolength{\textheight}{-12cm}


\begin{thebibliography}{99}

\bibitem{c1} P. Y. Pan, R. Qiao, L. Chen, et al., ``Agility meets stability: Versatile humanoid control with heterogeneous data," \emph{arXiv preprint arXiv:2511.17373}, 2025.

\bibitem{c2} D. Nootebos and A. J. Park, ``An Accessible STP-Based Framework for Autonomous Robot Soccer with Simple Robots," in \emph{2025 IEEE 16th Annual Ubiquitous Computing, Electronics \& Mobile Communication Conference (UEMCON)}, IEEE, 2025, pp. 0708-0716.

\bibitem{c3} M. S. Luiz, M. A. Pastrana, G. A. O. e Aguiar, et al., ``Behavior-Based Control with Learning from Demonstration for Path Following Applied to Mobile Robots Soccer," in \emph{2024 Latin American Robotics Symposium (LARS)}, IEEE, 2024, pp. 1-6.

\bibitem{c4} L. Yang, B. Werner, A. B. Ghansah, et al., ``Bracing for Impact: Robust Humanoid Push Recovery and Locomotion with Reduced Order Models," \emph{arXiv preprint arXiv:2505.11495}, 2025.

\bibitem{c5} A. Nagaraju, M. G. V. Kumar, Y. R. Devi, et al., ``Deep Reinforcement Learning for Low-Cost Humanoid Robot Soccer Players: Dynamic Skills and Efficient Transfer," in \emph{2023 Seventh International Conference on Image Information Processing (ICIIP)}, IEEE, 2023, pp. 316-320.

\bibitem{c6} M. Abreu, L. P. Reis, and N. Lau, ``Designing a skilled soccer team for robocup: Exploring skill-set-primitives through reinforcement learning," \emph{Neural Computing and Applications}, vol. 2025, pp. 1-36.

\bibitem{c7} H. Zhong, H. Zhu, and X. Li, ``Development of a simulation environment for robot soccer game with deep reinforcement learning and role assignment," in \emph{2023 WRC Symposium on Advanced Robotics and Automation (WRC SARA)}, IEEE, 2023, pp. 213-218.

\bibitem{c8} T. Feng, X. Wang, Y. G. Jiang, et al., ``Embodied AI: From LLMs to World Models [Feature]," \emph{IEEE Circuits and Systems Magazine}, vol. 25, no. 4, pp. 14-37, 2025.

\bibitem{c9} J. Hughes, A. Abdulali, R. Hashem, et al., ``Embodied artificial intelligence: Enabling the next intelligence revolution," in \emph{IOP Conference Series: Materials Science and Engineering}, IOP Publishing, 2022, vol. 1261, no. 1, p. 012001.

\bibitem{c10} H. Liu, D. Guo, and A. Cangelosi, ``Embodied intelligence: A synergy of morphology, action, perception and learning," \emph{ACM Computing Surveys}, vol. 57, no. 7, pp. 1-36, 2025.

\bibitem{c11} M. H. Nasiri, S. H. M. Zonouzi, and A. Salimi-Badr, ``Enhancing Decisions of Goalkeeper and Kicker Players in the RoboCup 2D Simulation League through Behavioral Cloning," in \emph{2024 20th CSI International Symposium on Artificial Intelligence and Signal Processing (AISP)}, IEEE, 2024, pp. 1-6.

\bibitem{c12} B. Santos, A. Cardoso, G. Leão, et al., ``Hierarchical Reinforcement Learning and Evolution Strategies for Cooperative Robotic Soccer," in \emph{2024 7th Iberian Robotics Conference (ROBOT)}, IEEE, 2024, pp. 1-6.

\bibitem{c13} T. Huang, J. Ren, H. Wang, et al., ``Learning humanoid standing-up control across diverse postures," \emph{arXiv preprint arXiv:2502.08378}, 2025.

\bibitem{c14} S. Xu, J. Liu, J. Tang, et al., ``Multi objective reinforcement learning driven task offloading algorithm for satellite edge computing networks," \emph{Scientific Reports}, vol. 15, no. 1, p. 24045, 2025.

\bibitem{c15} D. Affinita, F. Volpi, V. Spagnoli, et al., ``Multi-agent coordination for a partially observable and dynamic robot soccer environment with limited communication," \emph{arXiv preprint arXiv:2401.15026}, 2024.

\bibitem{c16} A. Taourirte and M. S. Mia, ``Multi-Agent Reinforcement Learning and Real-Time Decision-Making in Robotic Soccer for Virtual Environments," \emph{arXiv preprint arXiv:2512.03166}, 2025.

\bibitem{c17} T. M. Cao, H. A. Pham, M. Walter, et al., ``Multi-Agent Robot Swarms: A Review of Sensing and Perceptual Strategies for RoboCup Soccer," in \emph{2025 11th International Conference on Mechatronics and Robotics Engineering (ICMRE)}, IEEE, 2025, pp. 126-131.

\bibitem{c18} M. R. Ramadhan, A. W. Maulana, M. N. A. Atqiya, et al., ``Neural Network-Based Ball Trajectory Control of Solenoid Kickers for Autonomous Soccer Robots," in \emph{2025 International Seminar on Intelligent Technology and Its Applications (ISITIA)}, IEEE, 2025, pp. 70-75.

\bibitem{c19} T. P. Baldão, M. R. O. A. Maximo, and T. Yoneyama, ``Reinforcement Learning Applied to Very Small Size Soccer Decision-Making, Trajectory Planning and Control In Penalty Kicks," in \emph{2024 Brazilian Symposium on Robotics (SBR), and 2024 Workshop on Robotics in Education (WRE)}, IEEE, 2024, pp. 115-120.

\bibitem{c20} L. Ye, B. Xing, B. Liang, et al., ``Gewu Playground: an open-source robot simulation platform for embodied intelligence research," \emph{Science China Technological Sciences}, 2026, doi: 10.1007/s11431-025-3253-2.

\bibitem{c21} Z. Li, X. B. Peng, P. Abbeel, et al., ``Reinforcement learning for versatile, dynamic, and robust bipedal locomotion control," \emph{The International Journal of Robotics Research}, vol. 44, no. 5, pp. 840-888, 2025.

\bibitem{c22} A. Labiosa, Z. Wang, S. Agarwal, et al., ``Reinforcement learning within the classical robotics stack: A case study in robot soccer," in \emph{2025 IEEE International Conference on Robotics and Automation (ICRA)}, IEEE, 2025, pp. 14999-15006.

\bibitem{c23} Y. As, C. Qu, B. Unger, et al., ``SPiDR: A Simple Approach for Zero-Shot Safety in Sim-to-Real Transfer," \emph{arXiv preprint arXiv:2509.18648}, 2025.

\bibitem{c24} F. Lin, S. Huang, T. Pearce, et al., ``Tizero: Mastering multi-agent football with curriculum learning and self-play," \emph{arXiv preprint arXiv:2302.07515}, 2023.

\bibitem{c25} P. MacAlpine, M. Depinet, and P. Stone, ``UT Austin Villa 2014: RoboCup 3D simulation league champion via overlapping layered learning," \textit{Proceedings of the AAAI Conference on Artificial Intelligence}, vol. 29, no. 1, 2015.

\bibitem{c26} T. Haarnoja, B. Moran, G. Lever, et al., ``Learning agile soccer skills for a bipedal robot with deep reinforcement learning," \emph{Science Robotics}, vol. 9, no. 89, p. eadi8022, 2024.

\bibitem{c27} S. Li, Y. Pang, P. Bai, et al., ``Dynamic fall recovery control for legged robots via reinforcement learning," \textit{Biomimetics}, vol. 9, no. 4, p. 193, 2024.

\bibitem{c28} F. Shi, Y. Kojio, T. Makabe, et al., ``Reference-free learning bipedal motor skills via assistive force curricula," in \textit{Proc. The International Symposium of Robotics Research}, Cham: Springer Nature Switzerland, 2022, pp. 304-320.



\end{thebibliography}
\end{document}